\documentclass{article}

\PassOptionsToPackage{numbers, compress}{natbib}
\usepackage[preprint]{neurips_2021} %nonatbib
\usepackage{microtype}
\usepackage{graphicx}
\usepackage{subfigure}
\usepackage{booktabs} % for professional tables
\usepackage{multirow, makecell}

\usepackage[utf8]{inputenc} % allow utf-8 input
\usepackage[T1]{fontenc}    % use 8-bit T1 fonts
\usepackage{hyperref}       % hyperlinks
\usepackage{url}            % simple URL typesetting
\usepackage{amsfonts}       % blackboard math symbols
\usepackage{nicefrac}       % compact symbols for 1/2, etc.
\usepackage{xcolor}         % colors
\usepackage[symbol]{footmisc}

\newcommand{\wvu}{wav2vec-U 2.0}

\usepackage{amsmath}
\usepackage{amssymb}
\usepackage{mathtools}
\usepackage{amsthm}
\usepackage{pifont}

\usepackage[textsize=tiny]{todonotes}

\newcommand{\xdownarrow}[2][]{%
\left.{#1}\right\downarrow{#2}}

\title{Towards End-to-end Unsupervised Speech Recognition}

\author{%
  \stepcounter{footnote}Alexander H. Liu\thanks{Work done during an internship at Meta AI.} \quad Wei-Ning Hsu \textsuperscript{$\ddagger$} \quad Michael Auli \textsuperscript{$\ddagger$} \quad Alexei Baevski \textsuperscript{$\ddagger$}  \\
  \\
  \textsuperscript{$\dagger$} MIT CSAIL \textsuperscript{$\ddagger$} Meta AI \\
  \\
  \texttt{alexhliu@mit.edu} \\
  \texttt{\{wnhsu,abaeveski,michaelauli\}@fb.com} \\
}

\begin{document}

\maketitle

\begin{abstract}

Unsupervised speech recognition has shown great potential to make Automatic Speech Recognition (ASR) systems accessible to every language. However, existing methods still heavily rely on hand-crafted pre-processing. Similar to the trend of making supervised speech recognition end-to-end, we introduce \wvu~which does away with all audio-side pre-processing and improves accuracy through better architecture. In addition, we introduce an auxiliary self-supervised objective that ties model predictions back to the input.  Experiments show that \wvu~improves unsupervised recognition results across different languages while being conceptually simpler.
The code is available at \url{https://github.com/pytorch/fairseq/tree/master/examples/wav2vec/unsupervised}

\end{abstract}

\newcommand{\wvz}{wav2vec-U}
\newcommand{\wvpp}{wav2vec 2.0}
\newcommand{\wvppbase}{\textsc{Base}}
\newcommand{\wvpplarge}{\textsc{Large}}

\newcommand{\vox}{Libri-Light}
\newcommand{\libri}{Librispeech}
\newcommand{\libril}{Libri-light}
\newcommand{\voxsz}{LL-60k}
\newcommand{\librisz}{LS-960}
\newcommand{\libriunsz}{LS-860}
\newcommand{\devother}{\textsc{dev-other}}

% Symbols for spaces, sets and manifolds
\newcommand{\Inp}{\mathcal{X}}
\newcommand{\Feat}{\mathcal{Z}}
\newcommand{\QFeat}{\mathcal{Q}}
\newcommand{\Phonr}{\mathcal{P}^r}
\newcommand{\Phong}{\mathcal{P}^g}
\newcommand{\Context}{\mathcal{C}}
\newcommand{\Seg}{\mathcal{S}}

\newcommand{\R}{\mathbb{R}}

% % Macros
% \newcommand{\tsum}[3]{{\left[#1\right]}_{#2}^{#3}}

\newcommand{\E}{\mathop{\mathbb{E}}}

\newcommand{\y}{Y}
\newcommand{\x}{X}
\newcommand{\ze}{z}
\newcommand{\zq}{q}
\newcommand{\zqt}{\tilde{q}}
\newcommand{\s}{S}

\newcommand{\cc}{c}
\newcommand{\id}{i}
\newcommand{\sss}{s}
\newcommand{\p}{P}
\newcommand{\pr}{P^r}
\newcommand{\pbest}{\hat{\p}}
\newcommand{\pp}{p}
\newcommand{\prand}{\tilde{P}}

\newcommand{\gen}{$\mathcal{G}$}
\newcommand{\genm}{\mathcal{G}}
\newcommand{\dis}{$\mathcal{C}$}
\newcommand{\dism}{\mathcal{C}}

\newcommand{\e}{\mathbf{e}}
\newcommand{\Lmlm}{\mathcal{L}_{m}}
\newcommand{\Ld}{\mathcal{L}_{d}}
\newcommand{\Lf}{\mathcal{L}_{f}}

\renewcommand{\sectionautorefname}{\textsection} 
\renewcommand{\subsectionautorefname}{\textsection} 
\renewcommand{\subsubsectionautorefname}{\textsection} 

\newcommand{\Enot}[1]{\num[exponent-product = \times]{#1}}

\section{Introduction}
\label{intro}

State-of-the-art ASR models~\cite{gulati2020conformer,han2020contextnet} have greatly improved performance over the recent past.
These models have been relying heavily on human supervision - requiring thousands of hours of transcribed speech.
Consequently, the main limiting factor in building ASR systems for many more languages is the need for large amounts of labeled data, making it challenging or even impossible to build such supervised ASR systems for the vast majority of languages of the world~\cite{lewis2016ethnologue}.

Fortunately, recent advances~\cite{liu2018completely,yeh2018unsupervised,baevski2021unsupervised} have shown that it is possible to train ASR systems in an unsupervised manner where speech annotation is no longer required.
On a high level, these methods can be viewed as a two-step pipeline: a pre-processing step followed by the unsupervised training step, with the goal of transcribing speech into phoneme sequences.

The pre-processing step aims to transform input speech representations into features that are better aligned to the underlying phonetic content.
This is done through speech segmentation and feature engineering.
The goal of speech segmentation is to discover boundaries in the input sequence to mark the change from one phone to another.
In practice, the boundaries can be determined through clustering representations~\cite{baevski2021unsupervised} or other unsupervised methods~\cite{wang2018segmental,chen2019audio}.
Besides speech segmentation, prior works also considered feature engineering a necessary step.
From the input speech, they reduced the frequency by merging intra-segment features and the dimensionality through k-means clustering~\cite{liu2018completely} or principal component analysis (PCA)~\cite{baevski2021unsupervised}.

While the motivation of the pre-processing step is clear, the disadvantages are also obvious.
Prior work~\cite{chung2018unsup} discovered that the accuracy of segmentation can dramatically impact subsequent steps, yet unsupervised speech segmentation is usually far from perfect~\cite{kreuk2020self}.
This results in cascading errors from segmentation and other feature engineering steps.
To address these issues, we propose an end-to-end system, which, through a better architecture, is able to replace all audio-side pre-processing steps with a single speech-to-phoneme neural network.

In addition to relaxing the aforementioned limitation, we also improve the unsupervised ASR training step.
The core idea of this step is to train a model which maps features extracted from a model pre-trained on unlabeled speech data into phoneme sequences by restricting the output vocabulary distribution and n-gram structures to be similar to real phoneme sequences.
Existing solutions for imposing such restrictions include adversarial training~\cite{liu2018completely} and empirical output distribution matching~\cite{yeh2018unsupervised}.
However, these methods do not ensure that the output is closely related to the spoken content.
To make up for this deficiency, we propose a self-supervised auxiliary objective function that provides content-based guidance to the model by utilizing pseudo labels derived from the raw audio signal.

To summarize our work, we propose \wvu~ - an unsupervised ASR framework with an end-to-end style training and an improved objective function.
In our experiments, we validate the effectiveness of the pipeline simplification and the proposed modification of the objective function.
We show that \wvu~ performs better than the existing unsupervised frameworks across different languages from high-resource to low-resource.
We also combine \wvu~with self-training strategies~\cite{chen2019completely,baevski2021unsupervised} to compare against prior works and supervised systems. 
The implementation is based on fairseq~\cite{ott2019fairseq} and the code will be released soon.

\section{Background}
\label{background}

\subsection{Unsupervised ASR with Adversarial Training}

\citet{liu2018completely} first achieved unsupervised phoneme recognition on TIMIT~\cite{garofolo1993timit}, a small and clean English dataset for proof-of-concept experiments.
Their system takes MFCC representations of speech as the input and performs the pre-processing step combining speech segmentation with (a separately trained) segmental audio word2vec~\cite{wang2018segmental} and feature discretization with k-means clustering.
Using the segmented discrete features, they train a phone-based ASR system through adversarial training inspired by Generative Adversarial Network (GAN)~\cite{goodfellow2014generative}.
In the adversarial training framework, the ASR model serves as the generator which transcribes the discrete index sequence into phone sequence with a learnable embedding table followed by softmax activation.
A 2-layer convolution-based discriminator is trained to distinguish the output of the generator and real phone sequences.
The goal of the generator is to output phone sequences that cannot be distinguished by the discriminator.
By iteratively training the two modules, the generator can learn to map the pre-processed feature sequences into phone sequences in an unsupervised manner. 

\subsection{Wav2vec-U}
\label{subsec:w2vu_back}
Different from the prior work, \citet{baevski2021unsupervised} proposed wav2vec-U (Fig.~\ref{fig:w2vu}) that first achieved unsupervised speech recognition on benchmark speech recognition datasets in many different languages.
Wav2vec-U takes speech representations extracted from the wav2vec 2.0 model~\cite{baevski2020wav} as the input instead of human-defined features.
Such representations can be obtained by self-supervised training on audio and have been proven to be much more effective than human-defined features for ASR.

For the pre-processing step, wav2vec-U applies a sequence of feature engineering on the speech representations: k-means clustering input frames, reducing the representation dimension with PCA, merging adjacent PCA features that share the same k-means cluster, and finally, mean pooling features from adjacent pairs of timesteps to better match the length of the phone sequences.
To train the ASR model taking the pre-processed features as input, wav2vec-U followed the adversarial training approach with a single layer CNN generator and a two-layer discriminator.

Since the proposed \wvu~model is closely related to wav2vec-U, we detail the prior work in Section~\ref{subsec:w2vu} and highlight our contributions as follows: 1) we remove the need for the audio pre-processing step, and 2) we introduce an auxiliary objective that improves the segment transcriber.

\subsection{Self-training of Unsupervised ASR}
\label{subsec:st}
Though training unsupervised ASR through adversarial learning is feasible, results have shown that the performance of such models is usually far from supervised models.
To close the gap, \citet{chen2019completely} discovered that GAN training can be harmonized with Hidden Markov Models (HMMs), iteratively training GAN and HMM with pseudo labels provided by the previous step and this can improve the system performance by over 30\% after 3 iterations.
Following this method, wav2vec-U proposed a 3-stage pipeline, training GAN followed by HMM, followed by fine-tuning pre-trained wav2vec2.0 models.
For the HMM training, the unlabeled speech used for adversarial training is pseudo-labeled by the generator along with a phoneme-to-phoneme Weighted Finite-State Transducers (WFST; \citealt{mohri1997finite,mohri2002weighted}) to incorporate a phone-based language model for decoding.
Next, the unlabeled speech is re-labeled with the HMM using a phoneme-to-word WFST to incorporate a word-based language model for decoding.
Finally, the HMM-labeled data is used to fine-tune a pre-trained wav2vec2.0 model with the target being letters, decoded with a letter-to-word WFST.
The self-training strategy significantly boosted the performance of the unsupervised system, making it comparable to the state-of-the-art fully-supervised system from 2019~\cite{park2019specaugment}.
In this work, we followed the same self-training approach to compare \wvu~against existing ASR systems.

\section{Methodology}
\label{method}

\begin{figure}
\centering
\begin{minipage}{.4\textwidth}
  \centering
  \begin{center}
\centerline{\includegraphics[width=0.75\linewidth]{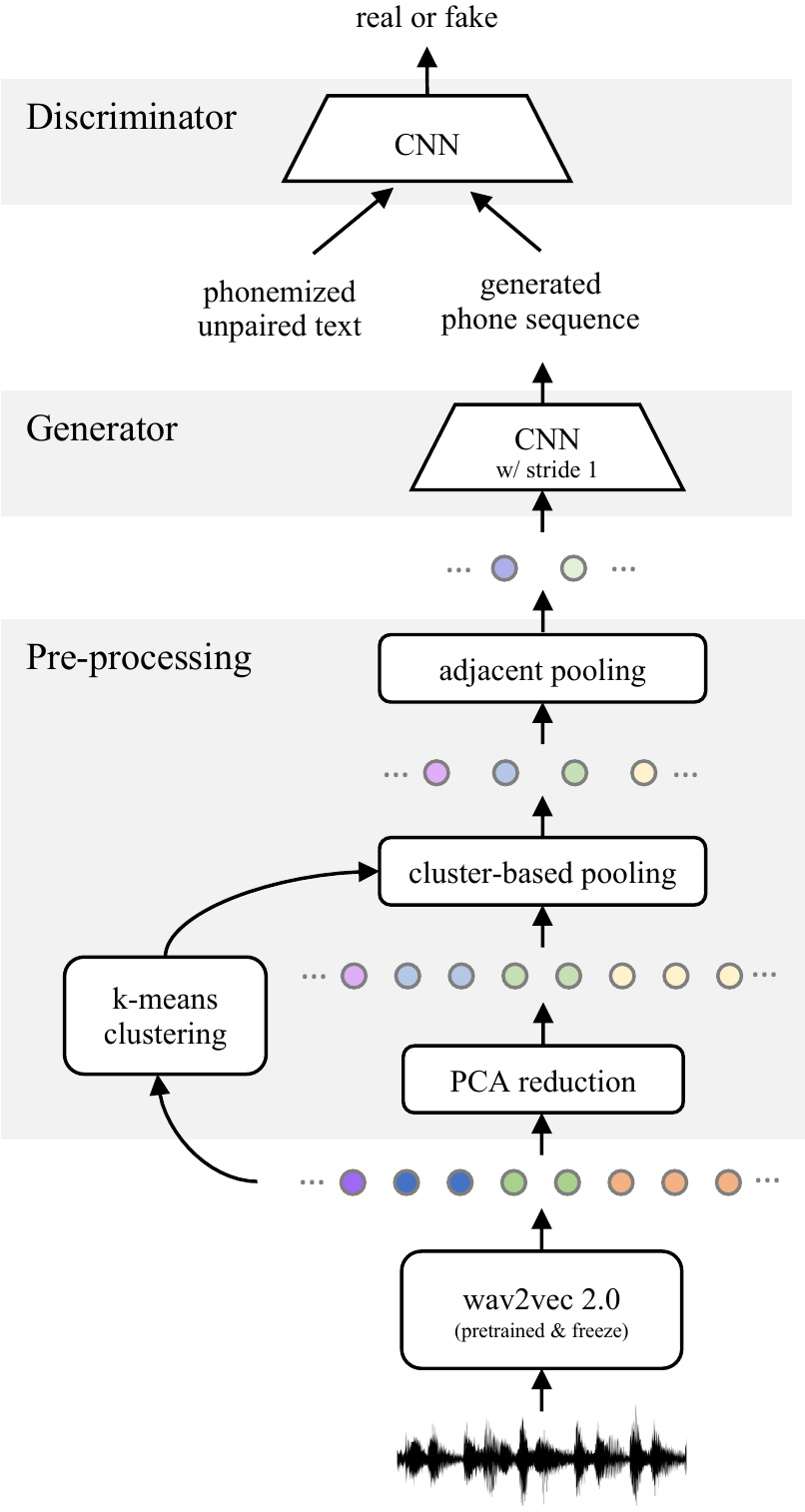}}
\end{center}
\caption{Wav2vec-U~\cite{baevski2021unsupervised}. The input wav2vec2.0 feature is pre-processed before feeding into the generator as described in Section~\ref{subsec:w2vu_back}. The generator is optimized through adversarial training against the discriminator as described in Section~\ref{subsec:w2vu}.}
% \vspace{-10pt}
\label{fig:w2vu}
\end{minipage}%
\hspace{0.05\textwidth}
\begin{minipage}{.5\textwidth}
\begin{center}
\centerline{\includegraphics[width=0.83\linewidth]{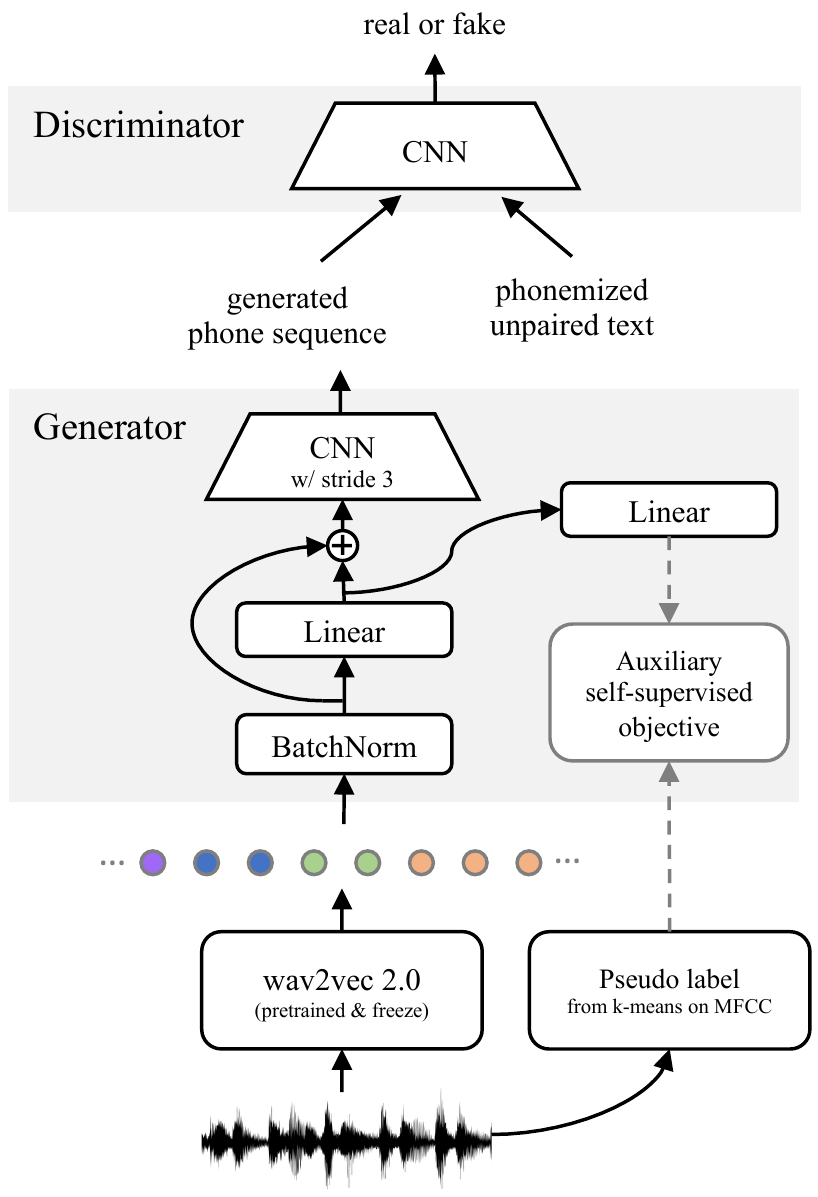}}
\end{center}
\vspace{-20pt}
  \caption{Proposed \wvu. The generator takes raw wav2vec 2.0 feature as input without pre-processing step as described in Section~\ref{subsec:simplify}. In addition to adversarial training, an auxiliary self-supervised objective is introduced with pseudo label derived from the raw waveform as described in Section~\ref{subsec:aux_loss}.}
% \vspace{-10pt}
\label{fig:overview}
\end{minipage}
\end{figure}

% \begin{figure*}[t]
% \begin{center}
% \centerline{\includegraphics[width=1.0\linewidth]{figures/w2vu.pdf}}
% \end{center}
% \caption{Wav2vec-U~\cite{baevski2021unsupervised}. The input wav2vec2.0 feature is pre-processed before feeding into the generator as described in Section~\ref{subsec:w2vu_back}. The generator is optimized through adversarial training against the discriminator as described in Section~\ref{subsec:w2vu}.}
% % \vspace{-10pt}
% \label{fig:w2vu}
% \end{figure*}
\subsection{Adversarial Training with wav2vec-U}
\label{subsec:w2vu}
To introduce the proposed \wvu, we begin with the adversarial training defined in wav2vec-U (Fig.~\ref{fig:w2vu}; \citealt{baevski2021unsupervised}).
Given $X$ the sequence of pre-processed features (described in Section~\ref{subsec:w2vu_back} and Fig.~\ref{fig:w2vu}) and $Y_\text{u}$ the phonemized unpaired text sequence, the goal of adversarial training is to train a Generator $\genm$ which transcribes $X$ into a phone sequence that cannot be distinguished from $Y_\text{u}$ by a discriminator $\dism$.
The full objective function is defined as
\begin{equation}
\label{eq:w2vu}
\begin{aligned}
    \min_{\genm}~\max_{\dism}~& 
\E_{Y_\text{u}} \left[ \log \dism(Y_\text{u}) \right] +
\E_{X} \left[ \log \left(1 - \dism (\genm(X)) \right) \right] \\
&- \lambda \mathcal{L}_{gp} + \gamma \mathcal{L}_{sp} + \eta \mathcal{L}_{pd} \\
\end{aligned}
\end{equation}
where the first two terms are the standard GAN objective \cite{goodfellow2014generative}; the third term
\begin{equation}
\begin{aligned}
    \mathcal{L}_{gp} &= \\
    &\E_{X,Y_\text{u},\alpha\sim U(0,1)} \left[ \left( \| \nabla \dism (\alpha~\genm(X) + (1-\alpha)~Y_\text{u} ) \| - 1 \right)^2 \right] \\
\end{aligned}
\end{equation}
is the gradient penalty on discriminator calculated with respect to a sample mixing real and fake input for better convergence~\cite{gulrajani2017improved} where $\alpha$ is the mixing weight sampled from uniform distribution; the fourth term
\begin{equation}
    \mathcal{L}_{sp} = \sum_{(\pp_t, \pp_{t+1}) \in \genm(X)} \| \pp_t - \pp_{t+1} \| ^2
\end{equation}
is the smoothness penalty to penalize locally inconsistent phone prediction from the generator where $\pp_t$ is the unnormalized generator output at time $t$; the fifth term
\begin{equation}
    \mathcal{L}_{pd} = \frac{1}{|B|} \sum_{\s \in B} -H_\genm(\genm(\s))
\end{equation}
is the phoneme diversity loss encouraging a diversed output over the phone set by maximizing entropy of the generator output distribution averaged over every frame in a mini-batch $B$; and $\lambda,\gamma,\eta$ are the weights for each term.

\subsection{Removing Audio Pre-processing}
\label{subsec:simplify}
As mentioned in Section~\ref{intro}, existing unsupervised ASR systems including wav2vec-U rely on hand-crafted pre-processing.
Interestingly, we discovered that the pre-processing step in wav2vec-U is not necessary.

Without the need of speech segmentation, a generator taking raw speech representation sequence as the input itself can be sufficient for unsupervised ASR with proper output frequency (i.e. number of phones predicted per second).
While there are various ways to control the output frequency of the generator, we found that simply changing the stride of the convolution neural network works well.
With the input wav2vec 2.0 features sampled at 50 Hz, we found that outputting a phone candidate at around 16 Hz by using a stride of 3 worked consistently well across all 9 languages tested in this work.
To further downsample the 16 Hz generator outputs to results closer to phone sequence, consecutive outputs sharing the identical most possible phone can be treated as a single output by random sampling as done in wav2vec-U.
In addition, we found that PCA-based dimensionality reduction can be replaced by a simple batch normalization~\cite{ioffe2015batch} along the time axis over the wav2vec 2.0 features.

As illustrated in Fig.~\ref{fig:overview}, \wvu~simply takes the raw representations extracted from wav2vec 2.0 model as input $X$ instead of pre-processed features as is done in wav2vec-U (Fig.~\ref{fig:w2vu}).
These modifications allow \wvu~to be trained without the need of a complicated pre-processing pipeline.
More importantly, learning the mapping from speech representations to phone sequences in an end-to-end style avoids the defects brought by hand-crafted pre-processing and performs better as demonstrated in Section~\ref{subsec:ablation}.

\subsection{Auxiliary Self-supervised Objective}
\label{subsec:aux_loss}
Since speech transcriptions are not available in the unsupervised setting, the generator relies only on adversarial training to learn the mapping between speech representations and phone units.
The generator can possibly learn to satisfy the adversarial criterion without learning the correct mapping - for example, by consistently producing the most common n-grams regardless of the input speech.

To address this issue, we introduce an auxiliary objective function that aids self-supervised learning by reconstructing a pseudo label sequence $Z=(z_1,z_2,...,z_T)$ derived from the input audio.
We use pseudo-labels generated by k-means clustering on Mel-frequency cepstral coefficient (MFCC) features with 64 clusters.
We found that this choice of pseudo labels works consistently well across different languages, and we show the effect of different choices of pseudo labels in Section~\ref{subsec:label}.
The auxiliary self-supervised objective for the generator is defined as
\begin{equation}
\mathcal{L}_{ss} = - \sum_{t} \log P_\genm(z_t|X)
\end{equation}
where $P_\genm(z_t|X)$, the probability of predicting the pseudo label $z_t$ at time $t$, is obtained by an additional linear transformation with the softmax function from the intermediate layer of the generator as illustrated in Fig.~\ref{fig:overview}.
The complete objective function of \wvu~can be obtained by adding the auxiliary objective to Eq.~\ref{eq:w2vu} with a weight $\delta$:
\begin{equation}
\label{eq:w2vu2}
\begin{aligned}
    \min_{\genm}~\max_{\dism}~& 
\E_{Y_\text{u}} \left[ \log \dism(Y_\text{u}) \right] -
\E_{X} \left[ \log \left(1 - \dism (\genm(X)) \right) \right] \\
&- \lambda \mathcal{L}_{gp} + \gamma \mathcal{L}_{sp} + \eta \mathcal{L}_{pd} + \delta \mathcal{L}_{ss} \\
\end{aligned}
\end{equation}

The benefit of introducing the auxiliary loss is two-fold: a) it provides a content-based regularization to ensure the output of the generator is closely related to the input speech and b) it guides the generator with a more explicit signal by approximating the underlying phone sequence with pseudo labels derived from speech.

\section{Experiments}
\label{experiment}
\subsection{Setup}

We follow \citet{baevski2021unsupervised} and perform experiments on 9 different languages from LibriSpeech~\cite{panayotov2015librispeech}, Multi-Lingual LibriSpeech~\cite{pratap2020mls}, and CommonVoice~\cite{ardila2019common}.
Unpaired text data is phonemized with an off-the-shelf phonemizer for English~\cite{g2pE2019} and other languages\footnote{\url{https://github.com/bootphon/phonemizer}} with silence token padded on both sides of each sentence and inserted at word boundaries with a probability of 0.5.
LJSpeech~\cite{ljspeech17} is also considered for preliminary experiments on end-to-end models outputting characters.
We summarize each dataset below and refer readers to prior work for more details on the dataset pre-processing.

\textbf{LibriSpeech \& Libri-Light.}
960 hours of English speech from LibriSpeech, discarding the transcriptions, is used for adversarial training.
In addition, 53.2k hours of speech from Libri-Light is used for self-training.          
For the unpaired text used to train \wvu~and language model, we used the official text corpus provided by LibriSpeech excluding the transcriptions of the speech data~\cite{synnaeve2020end}.
Evaluation is done on the standard dev/test sets.

\textbf{Multi-Lingual LibriSpeech.}
For each of the 6 languages (Dutch, French, German, Italian, Portuguese, and Spanish) selected from Multilingual LibriSpeech, 100 hours of speech is used for adversarial training and the official text corpus is used as unpaired text.
Evaluation is done on the standard test set.

\textbf{CommonVoice.}
We follow \citet{baevski2021unsupervised} to select 2 (Kyrgyz and Tatar) out of 38 languages from CommonVoice using 1.8 and 4.6 hours of unpaired speech respectively. For the unpaired text of both languages, we use a mixed text corpus following~\citet{baevski2021unsupervised}.
Evaluation is done on the dev/test set following~\citet{rivire2020unsupervised}.

\textbf{LJSpeech.}
This is a single-speaker corpus containing 24 hours of English speech.
Following prior works~\cite{ren2019almost,liu2020towards} on weakly-supervised speech recognition, we randomly select 300 utterances for validation and testing each and used the rest as unpaired speech for training.
For the unpaired text, we used the same data as the LibriSpeech setup.

\textbf{Implementation Details}
% \wn{the paragraphs here were quite fragmented so I tried merging them}
For English speech from LibriSpeech, the input features are extracted from the 15th layer of a pre-trained wav2vec2.0 Large model~\cite{baevski2020wav} encoded from the audio of which silences have been removed using an unsupervised voice activity detection tool~\cite{tan_rvad}.
For non-English speech, features are extracted from the XLSR-53~\cite{conneau2020unsupervised} model, a multi-lingual version of wav2vec 2.0.

As illustrated in Fig.~\ref{fig:overview}, the generator is a 2-layer CNN with batch normalization over the input features.
We found that initializing the scaling factor of batch normalization to 30 for wav2vec 2.0 Large and 35 for XLSR-53 features is critical for convergence.
During adversarial training, when obtaining the output probability distribution from the generator, consecutive frames that share the same most probable phone are merged into one by random selection.

The discriminator is a 2-layer CNN that uses the output of the generator or a one-hot representation of the real phone sequence as input.
A single scalar indicating whether the probability of the input sequence is real or not is obtained by taking the average over output sequence followed by the sigmoid activation. %\wn{is it sum or average? the scale would change wrt the sequence length if it's summation}

The generator and discriminator are trained for 100k steps in total interleaving the updates of each (50k updates for each) with a fixed learning rate of 5e-5 and 3e-4 respectively.
Each batch contains 160 audio samples and 160 randomly selected text samples.
With the choice of weights in Eq.~\ref{eq:w2vu2} being 1.0~/~1.5, 1.5~/~2.5, 0~/~3, 0.3~/~0.5 for $\lambda,\gamma,\eta,\delta$ respectively, we run each model with 4 different random seeds and we observe that over 80\% of the models will converge during training.
Models are selected with the Unsupervised Cross-Validation Metric proposed by \citet{baevski2021unsupervised}.

Pykaldi~\cite{can2018pykaldi} is used to decode the phone-based ASR outputs into words.
While the generator must be trained with a stride of 3, we found decoding using a language model with a stride of 2 (higher output frequency) leads to a more accurate output. 

\subsection{Comparison to wav2vec-U}
\label{subsec:ablation}

\begin{table*}[t]
\centering
\begin{center}
\caption{Interpolation from wav2vec-U (Fig.~\ref{fig:w2vu}) to \wvu~ (Fig.~\ref{fig:overview}). Phone Error Rate (PER) computed with greedy decoding on LibriSpeech \texttt{dev-other} set averaged over 8 runs. \textit{Freq.} refers to the frequency of sequence, i.e. number of tokens per second. 
\label{tab:ablation}}
\resizebox{\linewidth}{!}{
\begin{tabular}[t]{lccccccccc}
\toprule
 & \multicolumn{3}{c}{Pre-processing} & \multicolumn{4}{c}{Generator configuration} & \multicolumn{2}{c}{Result} \\
 \cmidrule(lr){2-4}\cmidrule(lr){5-8}\cmidrule(lr){9-10}
 & Adjacent  & Cluster & PCA & Batch & Linear & \multirow{1}{*}{Auxiliary} &  \multirow{2}{*}{Stride}  & \multirow{1}{*}{Freq.}  & Average \\
& pooling & pooling & reduction & norm. & proj. & loss &  & (Hz) & PER \\
\midrule
wav2vec-U    &  \checkmark & \checkmark & \checkmark &     -       &    -        &      -       & 1 & 14 & 18.8 $\pm$ 0.9 \\
\multirow{7}{*}{$\xdownarrow[\begin{gathered}
  \hfill \\
  \hfill \\
  \hfill \\
  \hfill \\
  \hfill \\
  \hfill \\
  \end{gathered}]{}$ } step (i)    &          -   & \checkmark & \checkmark &      -      &        -    &      -     & 1  & 28 & $>100$ \\
~~~~~~step (ii)       &         -    & \checkmark & \checkmark &       -     &      -      &     -      & 2 & 14 & 18.5 $\pm$ 0.6 \\
~~~~~~step (iii)      &        -     &        -    & \checkmark &     -       &       -     &    -       & 2 & 25 & $>100$ \\
~~~~~~step (iv)       &         -    &     -       & \checkmark &    -        &      -      &   -        & 3 & 16 & 19.0 $\pm$ 0.9 \\
~~~~~~step (v)        &     -        &      -      &      -      &      -      &      -      &    -       & 3 & 16 & $>100$ \\
~~~~~~step (vi)      &       -      &       -     &      -      & \checkmark &     -       &       -    & 3  & 16 & 16.4 $\pm$ 0.7 \\
~~~~~~step (vii)      &        -     &         -   &       -     & \checkmark & \checkmark &     -      & 3 & 16 & 15.9 $\pm$ 1.1 \\
\wvu~ &       -      &       -     &         -   & \checkmark & \checkmark & \checkmark  & 3 & 16 & \textbf{13.6 $\pm$ 0.9} \\
\midrule
\multicolumn{8}{l}{input wav2vec 2.0 feature} & 50  & - \\
\multicolumn{8}{l}{ground truth phone sequence} & $\sim$10  & - \\

\bottomrule
\end{tabular}
}
\end{center}
\vspace{-0.2cm}
\end{table*}
We first compare \wvu~with the closely related prior work wav2vec-U~\cite{baevski2021unsupervised} to show the benefit of removing pre-processing steps and introducing the auxiliary objective.

Table~\ref{tab:ablation} shows our proposed changes to wav2vec-U (Fig.~\ref{fig:w2vu} vs. Fig.\ref{fig:overview}) where changes are applied gradually from step (i) to step (vii).

Note, steps (i)-(iv) use pooling mechanisms to control the output frequency of the model. 
Removing these steps results in higher frequency and the model fails to converge to a reasonable accuracy. 
However, accuracy can be recovered by adjusting the output frequency by changing the stride of the generator (step (ii) \& (iv)). 
The fact that models with higher output frequency  (step (i) \& (iii)) failed to converge shows that having output frequency close to ground truth (around 10 Hz) is important.

We also observe a similar performance to the original framework after completely removing all pre-processing (step (iv)) and conclude that pre-processing speech representations in order to segment them into phone-like units are not necessary to obtain a working unsupervised ASR system, which can benefit from learning in an end-to-end fashion.

Comparing steps (iv) and (v), we see that some feature engineering in form of PCA dimensionality reduction is indeed important for adversarial training to work, but it can be replaced with a simple batch normalization layer as in step (vi).

Finally, looking at steps (vii) and the final step (viii), we see that introducing the auxiliary self-supervised objective improves PER significantly.

\subsection{Choice of Pseudo Labels for Self-supervision}
\label{subsec:label}
\begin{table}[h]
\centering
\caption{Different choice of pseudo label for  \wvu. Phone Error Rate (PER) computed with greedy decoding on LibriSpeech \texttt{dev-other} set averaged over 8 runs.
\label{tab:label}}
\begin{tabular}[t]{lccccc}
\toprule
Pseudo Label & \# of class & Avg. PER  \\
\midrule
\midrule
None & - & 15.9 $\pm$ 1.1 \\
\midrule
wav2vec2.0 VQ indices\textsuperscript{2} & 320$\times$2 & 16.6$\pm$2.2\\
\midrule
\multirowcell{3}[0ex][l]{k-means clustering\\wav2vec2.0 features\\} & 32 & 16.4$\pm$1.4\\
 & 64 & 15.5$\pm$1.8\\
 & 128 & 15.9$\pm$0.9\\
\midrule
\multirowcell{4}[0ex][l]{k-means clustering\\MFCC audio features} & 50 & 15.2$\pm$0.9\\
 & 64 & \textbf{13.6$\pm$0.9}\\
 & 100 & 14.8$\pm$1.3\\
 & 128 & 16.8$\pm$1.7\\
\bottomrule
\end{tabular}
\end{table}

Next, we investigate different options for pseudo labels that serve as an approximation of the underlying phone sequence for the auxiliary loss.
We experiment on LibriSpeech but found a similar trend on Multi-Lingual LibriSpeech and CommonVoice.

To acquire the pseudo label, we need to categorize input features according to some approximation of the underlying phone sequence from speech.
We first consider the wav2vec 2.0 vector quantization (VQ) indices\footnote{Indices from both codebooks in wav2vec2.0 are used as pseudo label with multi-task prediction} which come together with the input feature extraction.
We also consider k-means clustering on wav2vec 2.0 features which are used to detect phone boundary in wav2vec-U.
Finally, we consider performing k-mean clustering on the Mel-frequency cepstral coefficients (MFCC) audio features, which have been shown effective for self-supervised speech representation learning~\cite{hsu2020hubert}.

Note that the choice of the pseudo labels is important for the proposed objective.
Table~\ref{tab:label} shows the results.
VQ indices perform less well than the baseline method where no pseudo labels are used.
MFCC clustering yields the best results, leading to a reduction of up to 2.3\% PER, which is a 14\% relative improvement over the baseline.
The number of pseudo label classes can significantly affect results:
64 k-means clusters perform best on both wav2vec2.0 PCA features and the MFCC audio features.
Since the pseudo labels are extracted from audio, we believe a smaller label set might not be diverse enough to cover the phoneme inventory of the language and a larger label set will encode redundant information in the raw audio.

\subsection{Results on LibriSpeech}
\begin{table*}[t]
\centering
\begin{center}
\caption{
Word Error Rate (WER) on LibriSpeech with different language models (LM) on the standard LibriSpeech dev/test sets.
\label{tab:libri}
}
\resizebox{0.95\linewidth}{!}{
\begin{tabular}[b]{lccrrrr}
\toprule
\multirow{2}{*}{Model} & Unlabeled & \multirow{2}{*}{LM} & \multicolumn{2}{c}{dev} & \multicolumn{2}{c}{test} \\
\cline{4-5}\cline{6-7} 
{} & speech (hours) & {} & clean & other & clean & other \\
\midrule
\midrule
\multicolumn{7}{l}{\textbf{Supervised learning} w/ 960 hours of speech} \\
\midrule
~DeepSpeech 2~\citep{amodei2016deepspeech} & - & 5-gram & - & - & 5.33 & 13.25 \\
~Fully Conv~\citep{zeghidour2018w2l} & - & ConvLM & 3.08 & 9.94 & 3.26 & 10.47 \\
~TDNN+Kaldi~\citep{xu2018icassp} & - & 4-gram & 2.71 & 7.37 & 3.12 & 7.63 \\
~SpecAugment~\citep{park2019specaugment} & - & RNN & - & - & 2.5 & 5.8 \\
~ContextNet~\citep{han2020contextnet} & - & LSTM & 1.9 & 3.9 & 1.9 & 4.1 \\
~Conformer~\citep{gulati2020conformer} & - & LSTM & 2.1 & 4.3 & 1.9 & 3.9 \\
\midrule
\midrule
\multicolumn{7}{l}{\textbf{Semi-supervised learning} w/ 960 hours of speech}\\
\midrule
~Transf. + PL~\citep{synnaeve2020end} & 54k & CLM+Transf. &  2.00 & 3.65 & 2.09 & 4.11 \\
~IPL~\citep{xu2020iterative} & 54k & 4-gram+Transf. & 1.85 & 3.26 & 2.10 & 4.01 \\
~NST~\citep{park2020improved} & 54k & LSTM & 1.6 & 3.4 & 1.7 & 3.4 \\
~wav2vec 2.0~\citep{baevski2020wav} & 54k & Transf. & 1.6 & 3.0 & 1.8 & 3.3 \\
~wav2vec 2.0 + NST~\citep{zhang2020pushing} & 54k & LSTM & 1.3 & 2.6 & 1.4 & 2.6 \\
\midrule
\midrule
\multicolumn{7}{l}{\textbf{Unsupervised learning}} \\
\midrule
~wav2vec-U & 54k & 4-gram & 13.3 & 15.1 & 13.8 & 18.0 \\
~\wvu~ & 54k & 4-gram & 9.8 & 13.1 & 9.9 & 13.9 \\
%~\wvu~ + Self-Training & 960 & 4-gram & 3.8 & 6.4 & 4.2  & 7.0 \\
\midrule
\midrule
\multicolumn{7}{l}{\textbf{Unsupervised learning + Self-Training}} \\
\midrule
~wav2vec-U & 54k & 4-gram & 3.4 & 6.0 & 3.8  & 6.5 \\
~\wvu~ & 54k & 4-gram & 3.5 & 6.0 & 3.7  & 6.3 \\
\bottomrule
\end{tabular}
}
\end{center}
% \vspace{-0.4cm}
\end{table*}
\begin{table*}[t]
\centering
\begin{center}
% \scriptsize
\caption{Word Error Rate (WER) on the Multilingual Librispeech (MLS) for German (de), Dutch (nl), French (fr), Spanish (es), Italian (it) and Portuguese (pt).
\label{tab:mls}}
\resizebox{0.9\linewidth}{!}{
\begin{tabular}[b]{lcc|rrrrrr|r}
\toprule
\multirow{2}{*}{Model} & Labeled & \multirow{2}{*}{LM} & \multirow{2}{*}{de} & \multirow{2}{*}{nl} & \multirow{2}{*}{fr} & \multirow{2}{*}{es} & \multirow{2}{*}{it} & \multirow{2}{*}{pt} & \multirow{2}{*}{Avg.} \\
& data used & & & & & & & \\
\midrule
\midrule
\multicolumn{3}{l|}{Labeled training hours (full)} & 2k & 1.6k & 1.1k & 918 & 247 & 161  \\
\midrule
\midrule
\multicolumn{8}{l}{\textbf{Supervised learning}}\\
\midrule
~\citet{pratap2020mls} & full & 5-gram & 6.49 & 12.02 & 5.58 & 6.07 & 10.54 & 19.49 & 10.0 \\
\midrule
\midrule
\multicolumn{8}{l}{\textbf{Unsupervised learning}}\\
\midrule
~wav2vec-U & 0h & 4-gram & 32.5 & 40.2 & 39.8 & 33.3 & 58.1 & 59.8 & 43.9 \\
~\wvu~ & 0h & 4-gram & 23.5 & 35.1 & 35.7 & 25.8 & 46.9 & 48.5 & 35.9 \\
\midrule
\midrule
\multicolumn{7}{l}{\textbf{Unsupervised learning + self-training}} \\
\midrule
~wav2vec-U  & 0h & 4-gram & 11.8 & 21.4 & 14.7 & 11.3 & 26.3 & 26.3 & 18.6 \\
~\wvu~  & 0h & 4-gram & 11.5 & 17.6 & 12.8 & 10.9 & 18.6 & 20.6 & 15.3 \\
\bottomrule
\end{tabular}
}
\end{center}
% \vspace{-0.4cm}
\end{table*}

In Table~\ref{tab:libri}, we compare \wvu~with supervised, semi-supervised, and unsupervised prior work on the well-studied English benchmark dataset LibriSpeech.
Without self-training, \wvu~performs reducing word error rate (WER) by 2\% to 4\% compared to wav2vec-U.
This shows that removing the pre-processing steps and introducing the self-supervised auxiliary objective can lead to improved performance on speech recognition.
Next, we combine our method with the 3-stage self-training pipeline described in Section~\ref{subsec:st}.
Results show that WER can be significantly reduced with self-training, reaching below 3.7\% WER on the clean data and 6.3\% on the noisy data.

On the other hand, we observe the gap between wav2vec-U and \wvu~is much smaller after applying the self-training strategy.
This suggested that unsupervised ASR on LibriSpeech might be bottle-necked by the self-training pipeline.
Improving the self-training approach can likely further improve the performance of unsupervised ASR.

\subsection{Results on Multi-lingual LibriSpeech}

We have also tested our approach on a number of European languages from Multi-lingual LibriSpeech and show the results in Table~\ref{tab:mls}.

We observe that \wvu~also outperforms wav2vec-U in this setup, especially on the languages with higher error rate.
Overall, the proposed framework is able to improve the average WER on MLS by 8\% absolute without self-training.
This further demonstrates the benefit of having an end-to-end learning framework with self-supervision.

The results show that \wvu~consistently improves over wav2vec-U with self-training.
Average WER improves to 15.3\% compared to 18.6\% for wav2vec-U. 
This is a significant improvement since the supervised topline is 10\% WER.
On Portuguese, the WER is only 1.2\% worse than the supervised topline.

\subsection{Results on CommonVoice}

\begin{table}[t]

\centering
\begin{minipage}{.33\textwidth}
\centering
\caption{Phone Error Rate (PER) on CommonVoice.
\label{tab:cmv}}
\resizebox{\linewidth}{!}{
\begin{tabular}[t]{lcc}
\toprule
Model & tt & ky  \\
\midrule
\midrule
\multicolumn{3}{l}{\textbf{Supervised learning}}\\
\midrule
\citet{fer2017multilingually} & 42.5 & 38.7 \\
m-CPC~\citep{rivire2020unsupervised} & 42.0 & 41.2 \\
XLSR-53~\citep{conneau2020unsupervised} & 5.1 & 6.1 \\
\midrule
\midrule
\multicolumn{3}{l}{\textbf{Unsupervised learning}}\\
\midrule
wav2vec-U & 25.7 & 24.1 \\
\wvu~ & 22.5 & 24.3 \\
\midrule
\midrule
\multicolumn{3}{l}{\textbf{Unsupervised learning + HMM training}}\\
\midrule
wav2vec-U & 13.7 & 14.9 \\
\wvu~ & 10.1 & 15.2 \\
\bottomrule
\end{tabular}
}
\end{minipage}%
\hspace{0.01\textwidth}
\begin{minipage}{.65\textwidth}

\centering
\begin{center}
\caption{Recognition Error Rate on LJSpeech. Unit error rate refers to the greedy decoding phone error rate for phone-based models or character error rate for letter-based model. Word error rate is obtained through phone-to-word or letter-to-word decoding.
\label{tab:ltr}}

\resizebox{\linewidth}{!}{
\begin{tabular}[b]{lcc|cc}
\toprule
\multirow{2}{*}{Model} & Labeled & Phone-based & Unit  & Word  \\
& data used & (lexicon required) & error rate & error rate \\
\midrule
\midrule
\multicolumn{5}{l}{\textbf{Semi-supervised learning}}\\
\midrule
\multirow{2}{*}{\citet{ren2019almost}} & 20min & \checkmark & 11.7 & - \\
 & 10min & \checkmark & 64.2 & -  \\
\citet{liu2020towards} & 10min & \checkmark & 35.2 & -  \\
\midrule
\multicolumn{5}{l}{\textbf{Unsupervised learning}}\\
\midrule
\wvu~& 0min & \checkmark & 16.0 & 14.4 \\
\wvu~letter-based & 0min & \ding{55} & 34.6  & 64.0 \\
\bottomrule
\end{tabular}
}
\end{center}

\end{minipage}

\end{table}

Other than English and European languages, we experiment on two low-resource Turkic languages in CommonVoice and the results are shown in Table~\ref{tab:cmv}.
Since we only measure the phone error rate, we only use the HMM stage of the self-training pipeline.

Compared to the supervised topline using 17 hours of labeled speech, \wvu~is able to achieve 10.1\% / 15.2\% PER on Tatar / Kyrgyz using just 4.6 / 1.8 hours of speech. This once again shows the value of developing unsupervised ASR for low-resource languages.

\subsection{End-to-end Attempt on LJSpeech}

While \wvu~ removed the need for human-engineered pre-processing steps and pipeline design, it is still not fully end-to-end.
Like conventional supervised ASRs and existing unsupervised ASRs, \wvu~relied on a lexicon which is an expert-defined word-to-phone mapping. %, to produce the target of learning.
Since the phoneme inventory is defined by a set of distinctive sounds, learning a speech-to-phoneme transformation is naturally easier for the model than a speech-to-grapheme mapping or a speech-to-word mapping.
However, collecting a lexicon for every language can be difficult in practice as it requires expertise in both linguistics and phonetics.

We, therefore, evaluate \wvu~without relying on lexicon by replacing all phones (for both adversarial training and language modeling) with letters that can be directly inferred from the text data.
Results on LJSpeech with the end-to-end version of \wvu{} are presented in Table~\ref{tab:ltr}.
We first show that \wvu{}, using no labeled data at all, is already outperforming previous works on semi-supervised ASR~\cite{ren2019almost,liu2020towards} which use small amounts of human-labeled data.

When switching to an entirely letter-based system without a lexicon, the unit error rate increases substantially.
Such degradation is reasonable as the former needs to learn to spell alongside the mapping between acoustic and writing units.
Unfortunately, we also discover that the word error rate of the letter-based model is significantly higher than the phone-based model.
These results suggest that end-to-end unsupervised ASR could be possible but further research and development are necessary.

\section{Conclusion}

An end-to-end approach for unsupervised ASR is key to increasing applicability to low-resource languages.
In this work, we move towards this goal by removing the need for human-engineered pre-processing and by improving accuracy.
While \wvu{} is simpler and more accurate, additional work is required to enable a fully end-to-end approach.
Future work includes simplifying the self-training pipeline and removing the need for a phonemizer.
Possible solutions for the latter are incorporating unsupervised lexicon discovery~\cite{lee2015unsupervised} or training unsupervised ASR with graphemic text units such as letters directly as for supervised systems.

\newpage

{\small
\bibliography{main}
}

\end{document}